\documentclass[10pt,twocolumn,letterpaper]{article}

\usepackage{cvpr}
\usepackage{times}
\usepackage{epsfig}
\usepackage{graphicx}
\usepackage{amsmath}
\usepackage{amssymb}
\usepackage{multirow}
\usepackage{graphicx}

\usepackage{etoolbox}
\makeatletter
\patchcmd{\maketitle}
 {\def\@makefnmark}
 {\def\@makefnmark{}\def\useless@macro}
 {}{}
\makeatother

\usepackage[breaklinks=true,bookmarks=false]{hyperref}

\cvprfinalcopy 

\def\cvprPaperID{****} 

\begin{document}

\title{Baidu-UTS Submission to the EPIC-Kitchens Action Recognition Challenge 2019}

\author{ Xiaohan Wang$^{1,2}$, Yu Wu$^{1,2}$, Linchao Zhu$^2$, Yi Yang$^{1,2}$\thanks{This work was done when Xiaohan Wang and Yu Wu were interned at Baidu Research. 
Part of this work was done when Yi Yang was visiting Baidu Research during his Professional Experience Program.} \\
\texttt{\small \{xiaohan.wang-3,yu.wu-3,linchao.zhu\}@student.uts.edu.au; yi.yang@uts.edu.au}\\
$^1$Baidu Research, $^2$The ReLER lab, CAI, University of Technology Sydney
}

\maketitle

\begin{abstract}
In this report, we present the Baidu-UTS\footnote{This submission is a joint work by the ReLER lab at UTS and Baidu Research.} submission to the EPIC-Kitchens Action Recognition Challenge in CVPR 2019. This is the winning solution to this challenge.
 In this task, the goal is to predict verbs, nouns, and actions from the vocabulary for each video segment. The EPIC-Kitchens dataset contains various small objects, intense motion blur, and occlusions. It is challenging to locate and recognize the object that an actor interacts with.
To address these problems, we utilize object detection features to guide the training of 3D Convolutional Neural Networks (CNN), which can significantly improve the accuracy of noun prediction. Specifically, we introduce a Gated Feature Aggregator module to learn from the clip feature and the object feature. This module can strengthen the interaction between the two kinds of activations and avoid gradient exploding. 
Experimental results demonstrate our approach outperforms other methods on both seen and unseen test set.
\end{abstract}

\section{Introduction}
Egocentric (first-person) video analysis is an important task but less explored than third-person video understanding. It is valuable for practical applications such as human-computer interaction, intelligent wearable devices, and service robots. Due to the lack of sufficiently large datasets, the progress in this area has been relatively slow. 
Recently, a large-scale egocentric video dataset named EPIC-Kitchens \cite{Damen2018ScalingEV} has been released, which provides a new benchmark and has attracted much attention. 
The EPIC-Kitchens dataset is the largest dataset in first-person vision so far. It consists of 55 hours of recordings capturing all daily activities in the kitchens. The recognition task on the EPIC-Kitchens dataset is to predict the verb, noun, and the combination pair in each video segment. 

Egocentric action recognition is a challenging task.
Compared to third-person activity recognition, it requires to distinguish the object that human is interacting with from various small objects. The intense camera motion and occlusion make it more difficult to obtain an accurate prediction.
Therefore, direct adoption of algorithms like 3D CNN that work for third-person video recognition may not achieve promising results on this task.

To address this problem, we introduce an object detection model to extract more precise object-related features to guide the training of 3D CNN. 
Specifically, we extract the clip feature and the object feature using 3D CNN and Faster R-CNN \cite{Ren2015FasterRT}, respectively. 
Later, the two features are sent to a Gated Feature Aggregator module to produce a new representation for the final classification. 
This module can stabilize the training process and strengthen the interaction of the two different activations. To make the object feature more robust to the motion blur and occlusion, we feed the context frames between the center of the video clip to the detection model.
Our method outperforms the baseline models and achieves the state-of-the-art on the test sets.

\section{Related Work}

\begin{figure*}[t]
\center
\includegraphics[width=0.8\linewidth]{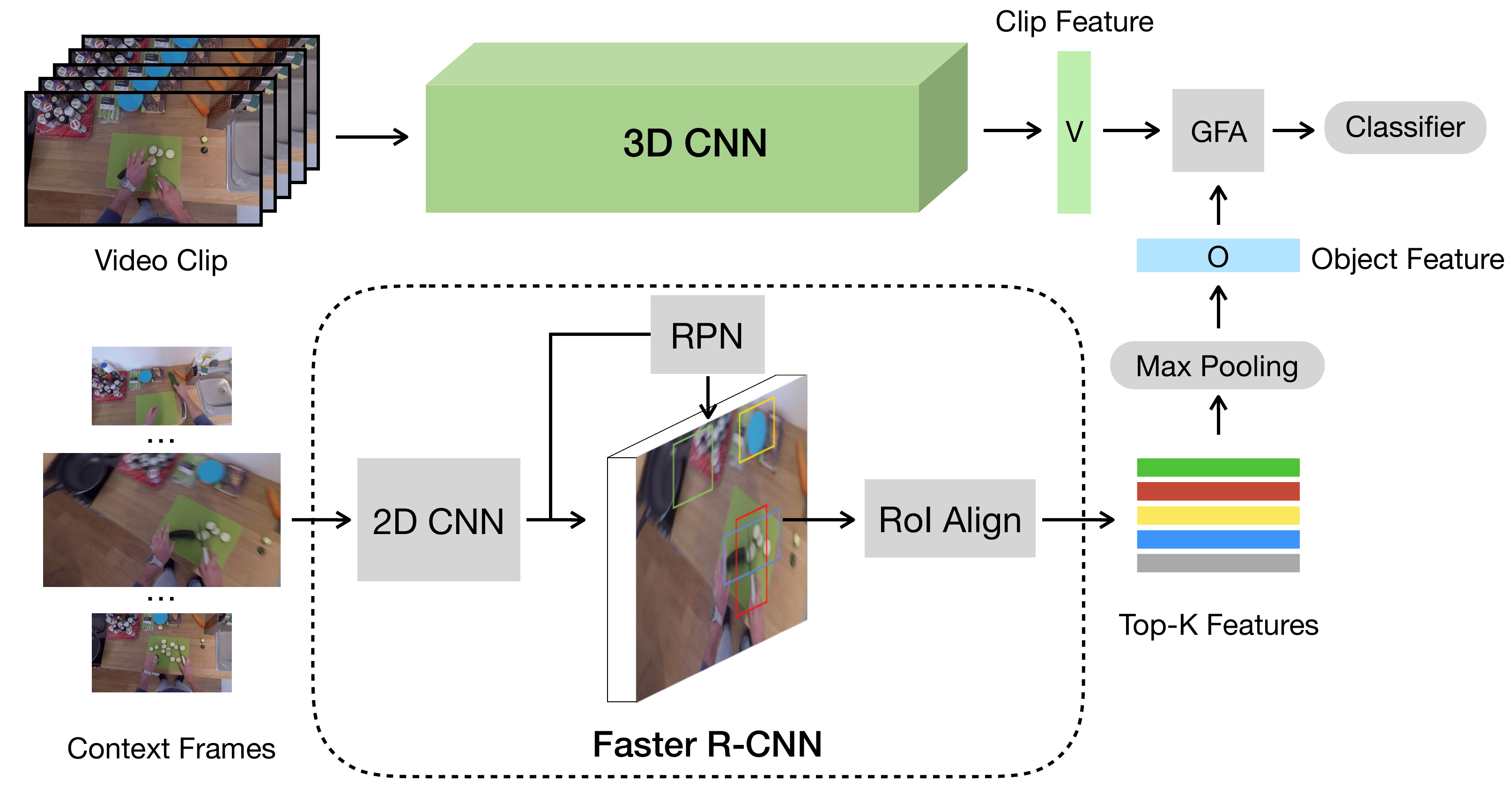}
\caption{
The overall framework of our approach.
}
\label{fig:framework}  
\end{figure*}

Third-person video classification has attracted lots of research works in the last a few years. Two-stream convolutional networks~\cite{simonyan2014two} utilize optical flow information for motion modeling, while 3D convolutional networks~\cite{tran2015learning,Carreira2017QuoVA,Xie2017AggregatedRT,zhu2019faster} recently achieved better performance than its 2D counterpart.
Recurrent Neural Networks (RNNs) are effective architectures for long  sequence modeling and have been found useful for video classification in \cite{Zhu_2017_CVPR,abu2016youtube}. Other aggregation methods like VLAD~\cite{xu2015discriminative}, actionVLAD~\cite{girdhar2017actionvlad} are also commonly used.

We discuss several methods evaluated on the EPIC-Kitchens dataset.
The authors of EPIC-Kitchens provide a baseline result on the recognition benchmark. They train Temporal Segment Network (TSN) \cite{TSN2016ECCV} to predict both verb and noun classes jointly. The two-stream TSN achieves the best performance on verb prediction and RGB-TSN outperforms their other models for noun prediction. However, without special design for egocentric videos, this state-of-the-art method for third-person video recognition does not achieve promising results, especially on noun classification.

The attention mechanism is efficient to locate the region of interest on the feature map. Sudhakaran et al. \cite{Sudhakaran_2019_CVPR} propose a Long Short-Term Attention model to focus on features from relevant spatial parts. They extend LSTM with a recurrent attention component and an output pooling component to track the discriminative area smoothly across the video sequence. Their model obtains a significant gain over the TSN baseline.

The object detection model is another powerful way to extract object-related features. Baradel et al. \cite{Baradel2018ObjectLV} propose to perform object-level visual reasoning about spatio-temporal interactions in videos through the integration of object detection networks. More recently, Wu et al. \cite{Wu2018LongTermFB} combine Long-Term Feature Banks that contains object-centric detection features with 3D CNN to improve the accuracy of noun recognition. 

According to the success of image recognition, pretraining on large scale dataset can boost the performance of deep learning models. Ghadiyaram et al. \cite{Ghadiyaram_2019_CVPR} construct a large-scale video dataset with verb-noun label space. They pretrain a deep 3D CNN on the data and then finetune the model on EPIC-Kitchens. Their model achieves relatively high results, especially on the unseen test set.

\section{Our Approach}

As shown in Fig.~\ref{fig:framework}, our framework consists of two branches. 
The first 3D CNN branch takes the sampled video clip as input and produces a clip feature. 
The second branch aims to extract the object-related features from the context frames. 
We sampled the frames within a window size $w$ at the center of the current clip. 
Then the pretrained object detector processes them frame by frame. We choose the top-K bounding boxes with the highest score and use RoIAlign \cite{He2017MaskR} to get the features from the feature maps of the 2D CNN. 
After that, the top-K features are max pooled and send to the Gated  Feature Aggregator (GFA) module with the clip feature. 
This module can guide the model to utilize the object-related information and find more discriminative channels. We describe the details of GFA in Sec. \ref{GFA}.  
The output of GFA is our final feature and is used to classify verbs and nouns.
\subsection{Base Models}
We use three 3D CNN backbones to extract video clip features. 
The first one is I3D \cite{Carreira2017QuoVA} which is proposed by Carreira and Zisserman. 
They inflate 2D CNN architectures to 3D and initialize the network with ImageNet \cite{Deng2009ImageNetAL} pretrained weights. 
We use the two-stream I3D for verb classification and RGB I3D for noun classification. We do not use optical flow inputs for noun as it does not contain enough information of object appearance. 
The other two backbones we used are 3D ResNet-50 \cite{Hara2018CanS3} and 3D ResNeXt-101\cite{Hara2018CanS3}. 
They have similar architectures as the 2D models, but the convolutional kernels are in 3D. 
All the three 3D CNN backbones are pretrained on the Kinetics-400 dataset \cite{Carreira2017QuoVA}.


 We use Faster R-CNN \cite{Ren2015FasterRT} pipeline to detect objects and extract object features. 
 The backbone of the detector is 2D ResNeXt-101 \cite{Xie2017AggregatedRT} with FPN \cite{Lin2017FeaturePN}, which is trained on 1600-class Visual Genome \cite{Krishna2016VisualGC,Anderson2018BottomUpAT} and then finetuned on EPIC-Kitchens \cite{Damen2018ScalingEV} detection dataset. 
 We train two detectors following the above steps. 
 One is $32\times8d$ ResNeXt-101 with 1024-dim output, and the other is $64\times4d$ ResNeXt-101 with 2048-dim output.
 
\subsection{Gated Feature Aggregator}\label{GFA}

\begin{figure}[t]
\center
\includegraphics[width=\columnwidth]{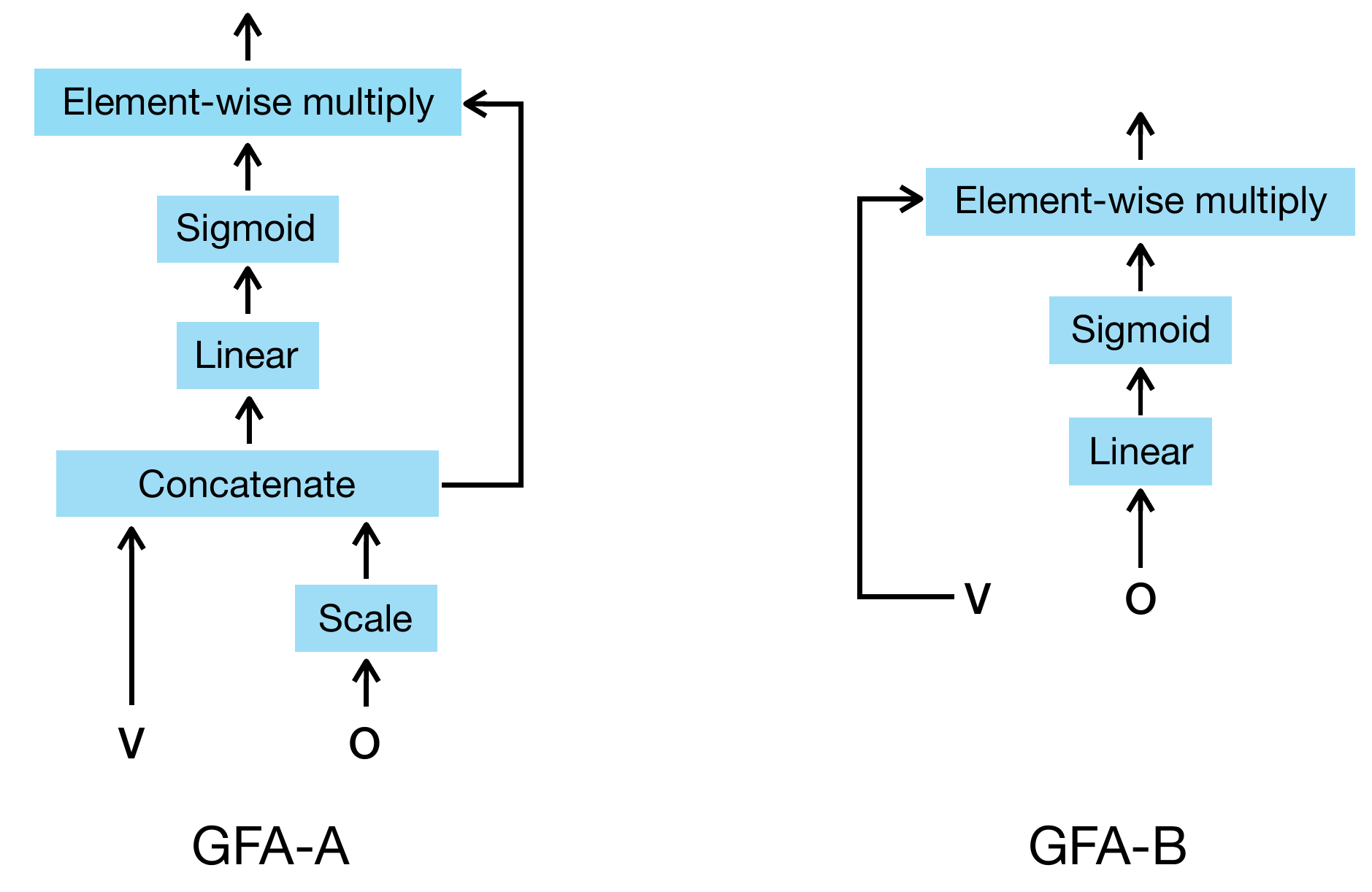}
\caption{
The two different types of GFA.
}
\label{fig:gfa}  
\end{figure}

Wu et al. \cite{Wu2018LongTermFB} propose to concatenate the object feature and the clip feature directly as the final representation. 
However, in our experiments, this method is sensitive to the backbones of 3D CNN and detector. 
When the two branches have different backbones, e.g., I3D and ResNext, the training loss is difficult to converge thus the final performance is not improved. 
To stabilize the training process and leverage the interdependencies of these two features, we design a Gated Feature Aggregator (GFA) module. 
As illustrated in Fig.\ref{fig:gfa}, GFA has two types.

\textbf{GFA-A}.
Since the amplitudes of the object feature $o$ and the clip feature $v$ might be different, we scale $o$ to enforce its amplitude to be approximate with $v$. 
The scaling operation can be performed by dividing a scalar. 
Another way of scaling is to multiply the $\ell_2$-normed $o$ by the amplitude of $v$. 
After that, the concatenated $o$ and $v$ is transformed to a new representation by self-gating mechanism \cite{Miech2017LearnablePW}. 
Formally, the output feature is computed as follows,
\begin{align}\label{eq:gfcA}
  F = \sigma(W[v,scale(o)]+b)\cdot[v,scale(o)],
\end{align}
where ``$[\ ]$'' indicates the concatenation operation. 
We have two motivations behind this design. 
First, we wish to avoid the gradient explosion by scale operation. 
Second, we want to strengthen the object-related channel using the gating operation. 

\textbf{GFA-B}. The instability of the training process is mainly caused by the concatenation operation. 
In this type, we multiply gated $o$ by $v$ in an element-wise manner instead of concatenation. 
The final representation $F$ is obtained as follows,
\begin{align}\label{eq:gfcB}
  F = \sigma(Wo+b)\cdot v.
\end{align}

\begin{table}[t]
\centering
\setlength{\tabcolsep}{3.3pt}
\begin{tabular}{|c|c|c|c|c|}
\hline
3D CNN   & Baseline & Concat & GFA \\ \hline
ResNet-50    & 25.07 & 29.74  & 32.99     \\ \hline
I3D RGB  & 27.92 & 23.97  &  34.13     \\ \hline
\end{tabular}
\vspace{2mm}
\caption{
Comparison of different models for noun recognition on the new train/val set (Top-1 accuracy).
}
\label{table:com}
\end{table}

\subsection{Action Re-weighting}

The actions are determined by the pairs of verb and noun. 
The basic method of obtaining the action score is to calculate the multiplication of verb probability and noun probability. 
However, there are thousands of combinations and most verb-noun pairs that do not exist in reality, e.g. ``open the knife''. In fact, there are only 149 action classes that have more than 50 samples \cite{Damen2018ScalingEV}. 
Following the approach in \cite{Wu2018LongTermFB}, we 
re-weight the final action probability by a prior, i.e.
\begin{align}\label{eq:rew}
  P(action=(v,n)) = \mu(v,n) P(verb=v)P(noun=n),
\end{align}
where $\mu$ is the occurrence frequency of action in training set. 

\section{Experiments}

\begin{table}[t]
\centering
\setlength{\tabcolsep}{3.3pt}
\begin{tabular}{|c|c|c|c|c|}
\hline
3D CNN       & Detector & GFA & top-1 & top-5 \\ \hline
ResNet-50    & - & - & 55.62  & 81.60     \\ \hline
ResNeXt-101  & - & - & 57.43  &  81.46     \\ \hline
I3D RGB      & - & - & 59.38  &  82.78     \\ \hline
I3D Flow     & - & - & 56.65      & 80.79      \\ \hline
I3D two-stream     & - & - & 61.44      & 83.60     \\ \hline
ResNet-50    & 1024 dim & Type A  & 57.61      &  82.64     \\ \hline \hline
Fusion    & - & - & 63.15     &  84.57     \\ \hline
\end{tabular}
\vspace{2mm}
\caption{
Performance of different models for verb recognition on the new train/val set.
}
\label{table:verb}
\end{table}

\begin{table}[t]
\centering
\setlength{\tabcolsep}{3.3pt}
\begin{tabular}{|c|c|c|c|c|}
\hline
3D CNN       & Detector & GFA & top-1 & top-5 \\ \hline
ResNet-50    & - & - & 25.07  & 46.84     \\ \hline
ResNeXt-101  & - & - & 25.68  &  46.52     \\ \hline
I3D RGB      & - & - & 27.92  &  52.85     \\ \hline
ResNet-50     & 1024 dim & Type A & 31.79      & 56.80      \\ \hline
ResNet-50    & 2048 dim & Type A & 32.99      &  57.81     \\ \hline 
ResNeXt-101     & 1024 dim & Type A & 30.79      & 56.69      \\ \hline 
I3D RGB & 1024 dim & Type B & 31.14 & 58.42 \\ \hline
I3D RGB & 2048 dim & Type B & 34.13 & 60.36 \\ \hline \hline
Fusion    & - & - &  39.09    &  65.00     \\ \hline
\end{tabular}
\vspace{2mm}
\caption{
Performance of different models for noun recognition on the new train/val set.
}
\label{table:noun}
\end{table}


\begin{table*}[]
\centering
\begin{tabular}{|c|c|c|c|c|c|c|c|c|}
\hline
\multirow{2}{*}{Model} & \multirow{2}{*}{data split} & \multirow{2}{*}{re-weighting} & \multicolumn{2}{c|}{verb} & \multicolumn{2}{c|}{noun} & \multicolumn{2}{c|}{action} \\ \cline{4-9} 
 &  &  & top-1 & top-5 & top-1 & top-5 & top-1 & top-5 \\ \hline
fused model & train/val & w/o & 63.15  &  84.57 & 39.09 &65.00  & 27.68  & 48.07 \\ \hline
fused model & train/val & w & 63.15 & 84.57 & 39.09 & 65.00 & 28.98 & 49.78 \\ \hline
fused model & trainval/test-s1 & w & 69.80  & 90.95  & 52.27  & 76.71  & 41.37  & 63.59  \\ \hline
fused model & trainval/test-s2 & w & 59.68  & 82.69  & 34.14  & 62.38  & 25.06  & 45.95  \\ \hline
\end{tabular}
\vspace{2mm}
\caption{Performance of the fused model on the train/val and trainval/test set.}
\label{tab:action}
\end{table*}
In all experiments, the inputs for 3D CNN are 64-frame video clips. 
The clips are randomly scaled and cropped to $224\times224$. 
We choose the top-10 bounding boxes of the context frames to extract object features. 
We adopt the stochastic gradient descent (SGD) with momentum 0.9 for model training.

We train our model for verb and noun independently. 
To validate our models, we split the training data to the new training and validation set following \cite{Baradel2018ObjectLV}.

For verb recognition, as shown in Table \ref{table:verb}, we train five different models on the new training set and evaluate their top-1 and top-5 accuracy on the validation set.
The two-stream I3D model (late fusion of I3D RGB and I3D flow) obtains the best performance, which can achieve 61.44\% top-1 accuracy and 83.60\% top-5 accuracy. 
Our ResNet-50 with GFA improves the top-1 accuracy by 1.99\% than the baseline model, where GFA is type A with the norm and scale operation.

For noun recognition, as shown in Table \ref{table:noun}, we experiment with eight models on the new train/val set. 
Due to the large margin improvement of the performance of noun recognition, we try more combinations of 3D CNN, detectors, and GFA. 
ResNet-50 with 2048-dim detection features and GFA-A results in 7.92\% improvement of top-1 accuracy. Besides, I3D RGB model with 2048-dim detection features and GFA-B achieves the highest top-1 accuracy at 34.13\%. As shown in Table. \ref{table:com}, the GFA module is more efficient than the direct concatenation.

For action recognition, as shown in Table \ref{tab:action}, we calculate the final action top-1 and top-5 accuracy of our fused model on train/val split in two ways. 
The re-weighting strategy improves the top-1 accuracy by 1.30\% and top-5 accuracy by 1.71\%. 

For the final submission, we train the above models on the whole training data. 
Our model ensemble achieves the best performance on both seen (s1) and unseen (s2) test set. 
The final results are shown in Table \ref{tab:action}.

\section{Discussion and Future Work}

In this paper, we report our method details for the EPIC-Kitchens action recognition task. We combine object features with the clip feature to predict the action.
To stabilize the training process and strengthen the interaction of different activations, we introduce a Gated Feature Aggregator, which has been validated to be important for feature learning.
Our model achieves the state-of-the-art on both seen and unseen test data. 

Our model is simple and does not introduce any interaction between the verb branch and the noun branch during training time. It can also be combined with the multi-modal information such as the narration of each segment following video-language models \cite{zhu2017uncovering}. It is also promising to adopt the model on action anticipation task owing to the more precise object features. 

{\small
\bibliographystyle{ieee_fullname}
\bibliography{egbib}
}

\end{document}